\documentclass[conference,10pt]{IEEEtran}
\usepackage{epsfig,epsf,rotating,setspace,latexsym,amsmath,amssymb,amsfonts,bm,theorem,subfigure,epstopdf}
\usepackage{cite,authblk}
\usepackage{bbm}
\usepackage{algorithm}
\usepackage[noend]{algpseudocode}
\usepackage{color}
\usepackage{mathtools}
\usepackage{soul}

\algrenewcommand\algorithmicforall{\textbf{foreach}}
\algrenewcommand\algorithmicindent{1em}

\definecolor{green1}{rgb}{0.2,0.7,0.2}

\algnewcommand\algorithmicforeach{\textbf{for each}}
\algdef{S}[FOR]{ForEach}[1]{\algorithmicforeach\ #1\ \algorithmicdo}

\IEEEoverridecommandlockouts
\allowdisplaybreaks

\begin{document}
 
\title{$r$Age-$k$: Communication-Efficient Federated Learning Using Age Factor}
 
\author{Matin Mortaheb \qquad Priyanka Kaswan \qquad Sennur Ulukus\\
        \normalsize Department of Electrical and Computer Engineering\\
        \normalsize University of Maryland, College Park, MD 20742\\
        \normalsize  \emph{mortaheb@umd.edu} \qquad \emph{pkaswan@umd.edu} \qquad \emph{ulukus@umd.edu}}
 
\maketitle

\begin{abstract}
Federated learning (FL) is a collaborative approach where multiple clients, coordinated by a parameter server (PS), train a unified machine-learning model. The approach, however, suffers from two key challenges: data heterogeneity and communication overhead. Data heterogeneity refers to inconsistencies in model training arising from heterogeneous data at different clients. Communication overhead arises from the large volumes of parameter updates exchanged between the PS and clients. Existing solutions typically address these challenges separately. This paper introduces a new communication-efficient algorithm that uses the age of information metric to simultaneously tackle both limitations of FL. We introduce age vectors at the PS, which keep track of how often the different model parameters are updated from the clients. The PS uses this to selectively request updates for specific gradient indices from each client. Further, the PS employs age vectors to identify clients with statistically similar data and group them into clusters. The PS combines the age vectors of the clustered clients to efficiently coordinate gradient index updates among clients within a cluster. We evaluate our approach using the MNIST and CIFAR10 datasets in highly non-i.i.d.~settings. The experimental results show that our proposed method can expedite training, surpassing other communication-efficient strategies in efficiency.
\end{abstract}

\section{Introduction}\label{sec:intro}
Federated learning (FL) \cite{mcmahan2017communication} is a distributed learning paradigm that enables multiple clients to collaboratively train a shared model while keeping the training data decentralized and private at the clients. The training with the clients is coordinated here by a centralized server known as the parameter server (PS). However, FL algorithms suffer from two major issues: data heterogeneity \cite{hetero_noniid_1, hetro_noniid_2} and communication overhead \cite{sattler2019robust, bernstein2018signsgd, wen2017terngrad, basu2019qsparse, Reisizadeh2020FedPAQAC, top_k, barnes2020rtop, stich2018sparsified, isik2023communication}. The statistical data heterogeneity stems from each client collecting its own data, with the local datasets in FL typically exhibiting different distributions, which can cause model divergence and degrade the FL performance. In addition, exchanging the hundreds of millions of parameters by clients with the PS in modern deep learning (DL) models introduces significant communication overhead on the clients. 

In this respect, we review the existing works and commonly used techniques that address these issues. To overcome the challenge of communication overhead, the gradients can be compressed using quantization \cite{Reisizadeh2020FedPAQAC, bernstein2018signsgd, wen2017terngrad, basu2019qsparse} or sparsification \cite{top_k, barnes2020rtop, stich2018sparsified, isik2023communication} techniques. The success of these compression techniques hinges on the sparse nature of gradient vectors intrinsic to neural networks (NN) during training. We take particular note of two compression strategies popular in the literature: \emph{top-k} sparsification \cite{top_k} and \emph{rTop-k} sparsification \cite{barnes2020rtop}. Under \emph{top-k} sparsification algorithm, clients transmit the indices with the top $k$ largest magnitudes from their gradient vector to the PS in each global iteration. Improving upon \emph{top-k}, \emph{rTop-k} sparsification algorithm entails clients first selecting $r$ $(r > k)$ indices with the largest magnitudes and then randomly picking a subset of $k$ indices from the selected indices to send to PS. The superior performance of \emph{rTop-k} over \emph{top-k} sparsification is attributed to the reduction in bias introduced by \emph{top-k} towards exploitation of only the significant entries of the gradient vector, thereby bringing in a strategic mix of exploration and exploitation of the gradient vector during training. 

On the other hand, data heterogeneity problem has been addressed in the literature with personalization \cite{arivazhagan2019federated, collins2021exploiting, FedGradNorm} and clustering \cite{morafah2022flis, mortaheb2023personalized, sattler2020clustered} methods. Personalization calls for the PS and the clients to initially train a common base model, with each client subsequently training a small header network tailored for its specific task. Through personalization, users can obtain distinct learning models that better fit their unique data, while still incorporating the common knowledge distilled from other devices’ data. Clustering, on the other hand, copes with data heterogeneity by grouping users with similar data distribution into clusters and training a specialized model for the clients in each cluster.

\begin{figure}[t]
    \vspace{0.1cm}
    \centerline{\includegraphics[width=1\linewidth]{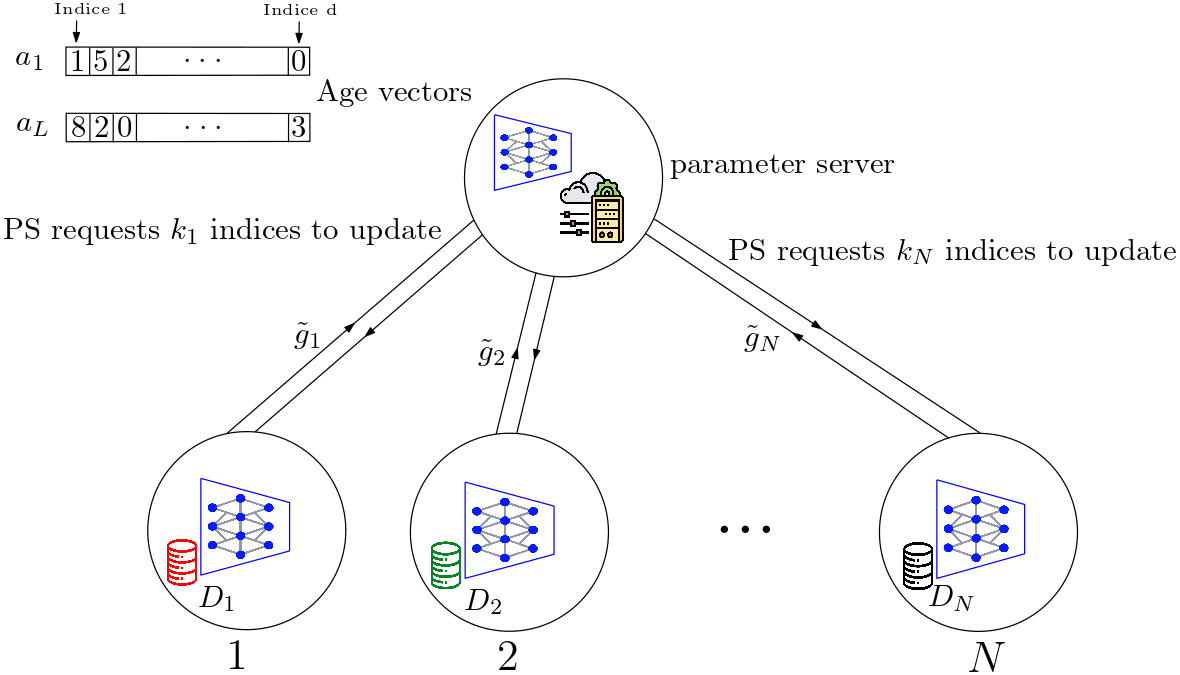}}
    \caption{System model for \emph{rAge-k} framework.}
    \label{system_model}
\end{figure}

In this work, we address these limitations, hitherto handled separately in the literature, simultaneously within a single comprehensive algorithm. We refer to this algorithm as \emph{rAge-k}, for it employs the age of information (AoI) metric to obtain sparsified gradient vectors for training the global model. AoI is defined as the time elapsed since the most recently received information was generated at its source, reflecting the staleness or outdatedness of information \cite{yates2021age}. In the context of \emph{rAge-k} sparsification, this information corresponds to the value in the gradient vector on a particular index, received from any client of a specific cluster, i.e., the source of information.  While functional aspects of AoI have been explored in distributed computing and learning \cite{yang2020age, ozfatura2020age, buyukates2020timely, wang2022age,  buyukates2021timely}, these works have no overlap with data heterogeneity or gradient sparsification based two-pronged objective of this work.  

The pivot of the \emph{rAge-k} algorithm is the introduction of age vectors at the PS for all the clusters. These age vectors have the same dimension as the number of parameters in the model, i.e., the dimension of the full gradient vectors, and aid in monitoring the age of gradient updates received on all the indices of the gradient vector. An age vector will not store the same age for all indices, since \emph{rAge-k}, being a sparsification based algorithm, involves communicating only a subset of indices in each round by any client to the PS, inducing nonuniform staleness with respect to different indices. With these age vectors, we can track the most and least frequently updated indices from each cluster, which serves several purposes. First, the age vectors help PS choose a small subset of indices to request updates on in each training round, thereby reducing the communication overhead of training. Second, the updating frequency of the indices tracked under this algorithm helps in identifying clients with statistically similar data, which can be grouped into clusters. The PS can exploit the cluster formations by merging the age vectors of these clients in the cluster, thereby enhancing the coordination of index updates within a cluster among clients. In scenarios where the PS is limited to requesting few updates from the clients, the merged vectors can be used by the PS to strategically choose a disjoint set of indices to request updates on from each individual client within the same cluster. 

The main contributions of our paper are as follows: 1) We introduce a novel FL framework that integrates the AoI metric to simultaneously address the two major limitations in FL, i.e., communication efficiency and data heterogeneity. 2) We utilize age vectors at the PS to strategically select subsets of indices in each training round to request gradient updates on from each client, and also to identify clients with statistically similar data and group them into clusters. 3) We conduct extensive experiments on our framework using the MNIST and CIFAR-10 datasets. The experimental results demonstrate the effectiveness of our framework in correctly clustering users with similar types of data, and achieving faster convergence and enhanced accuracy performance.

\section{System Model and Problem Formulation}
We consider an FL setting involving $N$ clients, where each client $i$ possesses a unique local dataset $D_i = \{(\mathbf{x}^{(i)}_j) \}_{j=1}^{n_i}$, with $n_i$ representing the size of the local dataset at client $i$. The clients aim to collaboratively train a model composed of $d$ parameters (with the $d$-dimensional parameter vector denoted by $\theta$) using their local datasets. The general FL objective with $N$ clients is expressed as,
\begin{align}
    \min_{\theta} \left\{ F(\theta) \triangleq \frac{1}{N} \sum_{i=1}^N \mathcal{L}_i(D_i,\theta_{i}^{t})\right\},
\end{align}
where $\mathcal{L}_i(D_i,\theta_{i}^{t})$ represents the loss specific to each client $i$ obtained by using their dataset $D_i$ with their local model parameters $\theta_{i}^{t}$ at time $t$.

In our framework, the parameter server maintains a $d$-dimensional age vector $a_l$ for each cluster $l \in [L]$ within the network, considering each client as an individual cluster upon initially joining the network of clients.

Each client performs $H$ local iterations before participating in a global iteration, i.e., sending its sparse gradients updates, composed of requested indices, to the PS. During global iterations, clients inform the PS of the indices in their gradient vector having the $r$ largest magnitudes. As mentioned earlier, the PS associates each client with a unique cluster, such that the number of clusters and the cluster constituents are redefined after every $M$ iterations. The PS uses the age vector associated with client-specific cluster to request updates from the client on a subset of $k_i$ ($k_i\ll d$) indices with the highest age out of the client-reported $r$ indices. Subsequently, clients create $\Tilde{g}_{i}^{t}$ by sparsifying their gradient vectors based on the requested $k_i$ indices, and transmit the non-zero elements of $\Tilde{g}_{i}^{t}$ and their index location to PS for a global model aggregation. The PS aggregates the received sparse gradient updates ($\Tilde{g}^{t} = \sum_{i=1}^{N} \Tilde{g}_{i}^{t}$) to update the global model $\theta$ and distributes the updated model back to all the clients. 

As shown in Algorithm~\ref{alg:rAge-k}, upon obtaining updates on the requested $k_i$ indices from each client $i$ by the PS, the age vector $a_l$, where $i \in$ cluster $l$, will get updated according to the following protocol, 
\begin{align} \label{spectral}
    a_l^{t}(j)= 
    \begin{cases}
        0, & \text{if } j \in [k_l],\\
        a_l^{t-1}(j) + 1, & \text{otherwise},
\end{cases}
\end{align}
where $a_l^{t}(j)$ is age of cluster $l$ at time $t$ corresponding to index $j$. In other words, the age of the requested indices are reset to $0$ since they got updated recently, however the age of the remaining indices increase by one, indicating that they are further outdated at the PS.

At intervals of every $M$ iterations, upon receiving the sparsified gradient updates from the clients, the PS initiates an algorithm to identify clients with similar types of data, aiming to group them into the same cluster.  We associate a frequency vector $f^t[i]$ of size $d$ with client $i$, in which the coordinate $j$ indicates how many times the index $j$ was requested from client $i$ by PS until time $t$. The PS carries out the clustering process by calculating a similarity matrix for each pair of clients from the dot product of their frequency vectors as follows,
\begin{align} \label{eq:distance_calculation}
    d^t[i_1,i_2] = \frac{\langle f^t[i_1],f^t[i_2] \rangle}{\langle f^t[i_1],f^t[i_1] \rangle}.
\end{align}
Next, the PS employs the DBSCAN (density-based spatial clustering of applications with noise) method \cite{ester1996density} to cluster clients based on the similarity matrix computed earlier. DBSCAN distinguishes itself from its counterpart clustering methods in that it does not require the number of clusters to be predetermined. DBSCAN relies on a density-based notion of clustering to group data points, where it identifies clusters as areas of high density of points separated by areas of low density. DBSCAN takes in two parameters as input: \emph{eps}, the maximum distance between two points for them to be considered in the same neighborhood, and \emph{minPts}, the minimum number of points required to form a dense region. Points inside the dense regions are labelled as core points, while the points close to dense regions but not part of them are labelled as border points, while all other points are considered noise. This method is adept at discovering clusters of arbitrary shapes and is robust to outliers. Note that when a client is added to an existing cluster, its age vector is merged with that of the cluster. Conversely, if DBSCAN determines that a client significantly deviates from its current cluster and the client is reassigned to a different group, the age vector relevant for that client is automatically reset due to the changed cluster identity. The overall training process and the age-function are detailed in Algorithm~\ref{alg:first_method} and Algorithm~\ref{alg:rAge-k}, respectively.

\begin{algorithm}[h]
    \caption{Training with our proposed algorithm}
    \label{alg:first_method}
    \begin{algorithmic}[1]
        \State {\bfseries Input:} step sizes $\eta$, initialization $\theta$, $H$, $k$, $r$.
        \For{t = 1, \ldots, $T$}
        \ForEach {$i \in N $ (in parallel)} \Comment{Local Iteration}
        \State Compute client-loss $\mathcal{L}_i(D_i,\theta_{i}^{t})$
        \State $\theta_{i}^{t+1} \leftarrow \theta_{i}^{t} - \eta \nabla_{\theta_i}\mathcal{L}_i(D_i,\theta_{i}^{t})$
        \EndFor
        \If{$t \% H = 0$} \Comment{Global Iteration}
        \State $\Tilde{g}_{i}^{t}$ = rAge-k$(\nabla_{\theta_i}\mathcal{L}_i(D_i,\theta_{i}^{t}), a_i, k, r)$
        \State Client $i$ sends non-zero elements of $\Tilde{g}_{i}^{t}$ and its location to PS
        \EndIf
        \State \textbf{PS:} Update the age vector
        \State \textbf{PS:} $\Tilde{g}^{t} = \sum_{i=0}^{N} \Tilde{g}_{i}^{t}$
        \State \textbf{PS:} Update global model $\theta^t$ based on $\Tilde{g}^{t}$
        \State Send global model $\theta^t$ to all clients
        \If{$t \% M = 0$} \Comment{Cluster Checking at PS}
        \ForEach{$i,j \in N $ (in parallel)}
        \State Calculating $d^t[i,j]$ using (\ref{eq:distance_calculation}) to form D
        \EndFor
        \State Clustering = DBSCAN(\emph{eps},\emph{minPts}).fit($D$)
        \EndIf
        \EndFor
    \end{algorithmic}
\end{algorithm}

\begin{algorithm}[h]
    \caption{\emph{rAge-k} function}
    \label{alg:rAge-k}
    \begin{algorithmic}[1]
        \State {\bfseries rAge-k} (gradient-vector($g$), age-vector($a$), $k$, $r$):
        \State\hspace{\algorithmicindent} $\Tilde{g}$ = Zeros($g$.shape)
        \State\hspace{\algorithmicindent} (Top-val, Top-ind) = topk(abs($g$),r)
        \State\hspace{\algorithmicindent} (Topage-val, Topage-ind) = topk(age[Top-ind],k)
        \State\hspace{\algorithmicindent} Top-ind = Top-ind[Topage-ind]
        \State\hspace{\algorithmicindent} $\Tilde{g}$[Top-ind] = $g$[Top-ind]
        \State\hspace{\algorithmicindent} $a$ += 1
        \State\hspace{\algorithmicindent} $a$[Top-ind] = 0
        \State\hspace{\algorithmicindent} {\bfseries return} $\Tilde{g}$, Top-ind, $a$.
    \end{algorithmic}
\end{algorithm}

\subsection{Convergence of rAge-k}
Define  $f^{(i)}(\theta)=\mathbb{E}_{X \sim \mathcal{D}_i}\left[\mathcal{L}_i\left(X; \theta\right)\right]$. We make the following assumptions:
\begin{enumerate}
    \item Smoothness: The loss functions $f^{(i)}$ are $L$-smooth:
    \begin{align}
        f^{(i)}(\theta) \leq &f^{(i)}(\theta')+\left\langle\nabla f^{(i)}(\theta'), \theta-\theta'\right\rangle\nonumber\\
        &+\frac{L}{2}\left\|\theta-\theta'\right\|_2^2,  \quad \text{for any pair $\theta, \theta'$}.
    \end{align} 
    \item Bounded second moment of gradients: For some constant $0 \leq G<\infty$, we have
    \begin{align}
        \mathbb{E}_{X \sim \mathcal{D}_i}\left\|\nabla \mathcal{L}\left(X; \theta\right)\right\|_2^2 \leq G^2.
    \end{align}
    \item The ratio of the magnitudes of the largest magnitude index to the $r$th largest magnitude index is bounded by $\beta$ at all clients in all iterations.
\end{enumerate}

\textit{Definition (Compression operator)}: A (potentially randomized) function  ${Comp }_k$ is a compression operator if there exists a constant $\gamma \in(0,1]$ (depending on sparsification parameter $k$ and gradient dimension $d$) such that
\begin{align}
    \mathbb{E}\left\|\theta-{Comp}_k(\theta)\right\|_2^2 \leq(1-\gamma)\|\theta\|_2^2.
\end{align}

By Assumption~$3$, \emph{rAge-k} can be shown to be a compression operator with $\gamma=\frac{k}{k+(r-k)\beta+(d-r)}$, and \emph{rAge-k} therefore enjoys the convergence guarantees of $O(\frac{1}{\sqrt{T}})$ from \cite[Thm.~1]{basu2019qsparse}. Note that when $k=r$, $\gamma$ becomes simply $\frac{k}{d}$. Loosening $r$ increases $\beta$, hence higher exploration of gradient indices in our framework comes with less strict convergence guarantees.

\section{Experimental Results} \label{sec:Exp_section}

\subsection{Dataset Specifications}
In our experiments, we utilize two datasets: I) MNIST dataset comprising 28 × 28 gray-scale images of handwritten digits categorized into 10 classes, with the training (test) set of 60,000 (10,000) images, II) CIFAR10 dataset comprising 32 x 32 RGB images divided into 10 classes, each representing a specific object category such as airplanes, automobiles, birds, cats, deer, dogs, frogs, horses, ships, and trucks, with the training (test) set of 50,000 (10,000) images.

\begin{table}[h!]
    \caption{Network Model}
    \label{network_model}
    \vskip 0.1in
    \begin{center}
    \begin{small}
    \begin{sc}
    \begin{tabular}{lcccr}
    \hline
    %\abovespace\belowspace
    Network 1 & network 2 \\
    Total par (39760) & Total par (2515338) \\
    \hline
    %\abovespace
    FC(784,50)    & Conv2d(3, 64, 3)+BN(64)\\
    Relu  & MaxPool2d(2, 2)\\
    FC(50,10)  & Conv2d(64, 128, 3)+BN(64)\\
    Softmax    & Conv2d(128, 256, 3)+BN(256)\\
      & Conv2d(256, 512, 3)+BN(512)\\
       & FC(2048, 128)\\
       & FC(128, 256)\\
       & FC(256, 512)\\
       & FC(512, 1024)\\
       & FC(1024, 10)\\
    \hline
    \end{tabular}
    \end{sc}
    \end{small}
    \end{center}
    \vskip -0.1in
\end{table}

\subsection{Model and Hyperparameters}
We use Adam optimizer with learning rate $ \eta = 1 \times 10^{-4}$ to train the client models for both datasets. The model architectures for the MNIST and CIFAR10 datasets are detailed in Table~\ref{network_model}. For MNIST dataset, the model contains 39,760 parameters, and we set $r=75$ and $k=10$. We perform $H=4$ local iterations before a global iteration wherein clients send their sparse gradient updates to the PS. The PS updates the clusters in the network every $M=20$ iterations. For CIFAR10 dataset, the model contains 2,515,338 parameters, and we set $r=2500$ and $k=100$. We perform $H=100$ local iterations before the global iteration, and the PS updates the clusters every $M=200$ iterations. For both MNIST and CIFAR10 datasets, we consider batch size to be 256 during training.

\subsection{System Performance}

With MNIST dataset, to create a highly non-i.i.d.~scenario and evaluate the effectiveness of DBSCAN in clustering clients with similar data, we structure our distributed learning system with $10$ clients. Each client contains the images from two categories, and further, there are five pairs of clients such that the clients of every pair  have data from the same categories. Specifically, the local datasets of clients $1$ and $2$ come from the same categories/distribution as both handle images with label $0$ and $1$, likewise clients $3$ and $4$ handle images with label $2$ and $3$, and so on.
 
\begin{figure}[]
 	\begin{center}
 	\subfigure[]{%
 	\includegraphics[width=0.49\linewidth]{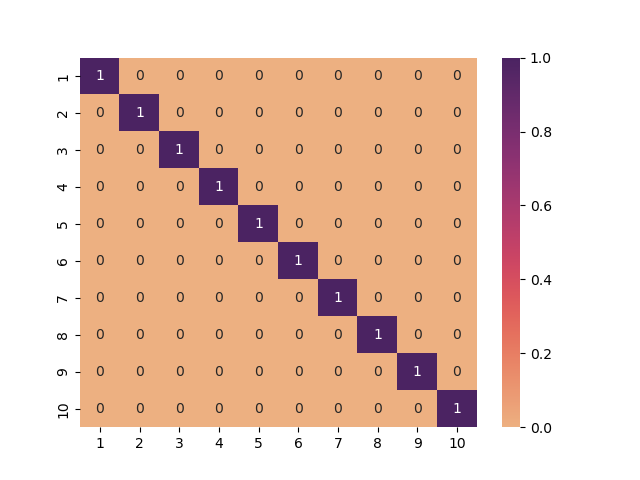}}
 	\subfigure[]{%
 	\includegraphics[width=0.49\linewidth]{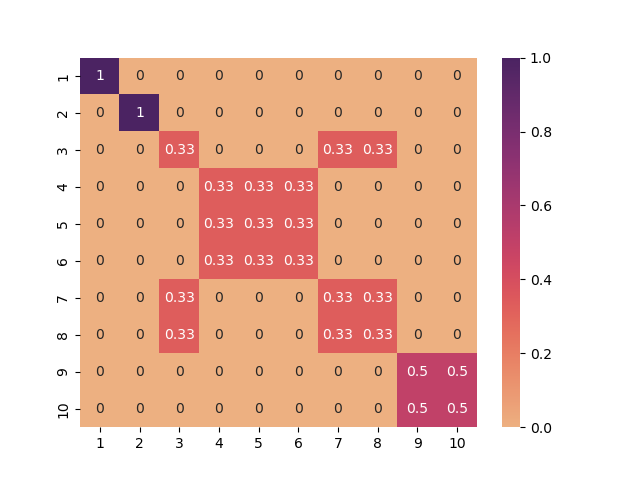}}\\ 
 	\subfigure[]{%
 	\includegraphics[width=0.49\linewidth]{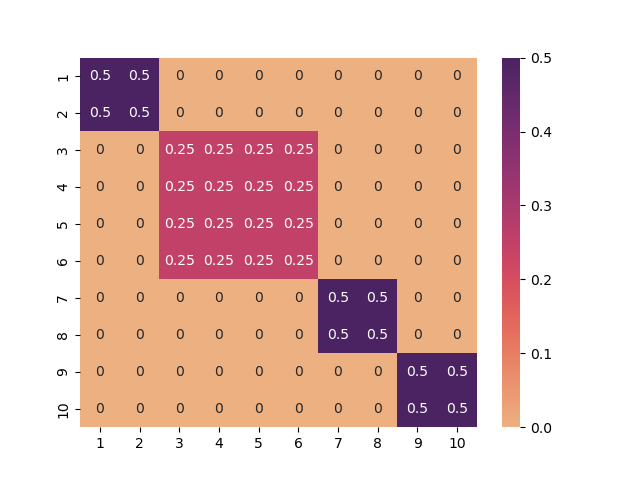}}
 	\subfigure[]{%
 	\includegraphics[width=0.49\linewidth]{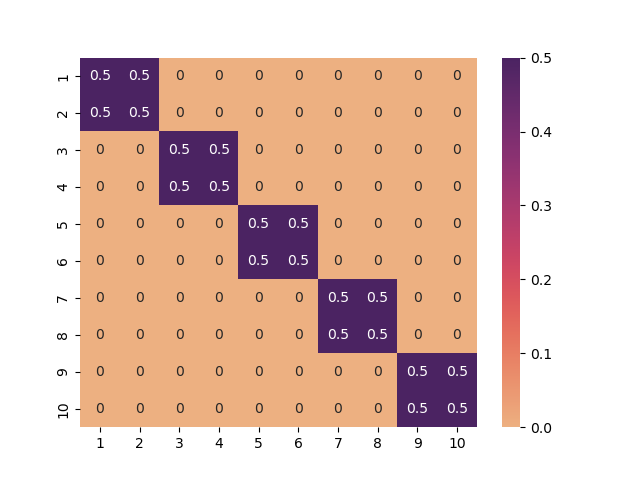}}
 	\end{center}
 	\caption{Heatmap of the connectivity matrix achieved by the DBSCAN method over different epochs, 
        (a) iteration 1, (b) iteration 21, (c) iteration 41, (d) iteration 61.}
 	\label{heatmap_synthetic}
\end{figure}

The heatmap of the weight assigned to pairs of clients inside the connectivity matrix at epochs $1$, $21$, $41$, $61$, shown in Fig.~\ref{heatmap_synthetic}, illustrates the effectiveness of DBSCAN in identifying users with similar image categories/data distributions. The evolution of heatmap in Fig.~\ref{heatmap_synthetic} demonstrates that the PS successfully detects similar clients, such that the clustering remains broadly the same until the end of the training.

To test the effectiveness of our communication-efficient framework in terms of accuracy performance, we conduct a comparative analysis with \emph{rTop-k}, a popular communication-efficient algorithm. In \emph{rTop-k}, the client first selects the $r$ indices with the largest magnitudes in its gradient vector and randomly picks $k$ indices from the selection to send updates on to the PS. The aggregation mechanism at the PS in \emph{rTop-k} remains consistent with our framework. For a meaningful comparison of \emph{rAge-k} with \emph{rTop-k}, in our simulations we use the same $r$ and $k$ values in both algorithms.

\begin{figure}[!h]
    \begin{center}
    \subfigure[]{%
    \includegraphics[width=0.49\linewidth]{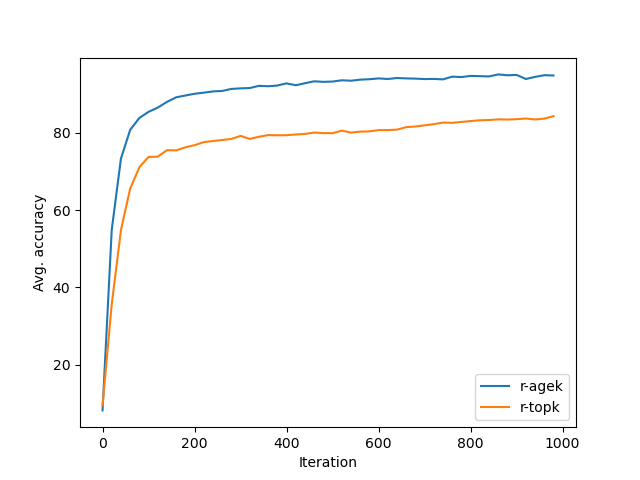}}
    \subfigure[]{%
    \includegraphics[width=0.49\linewidth]{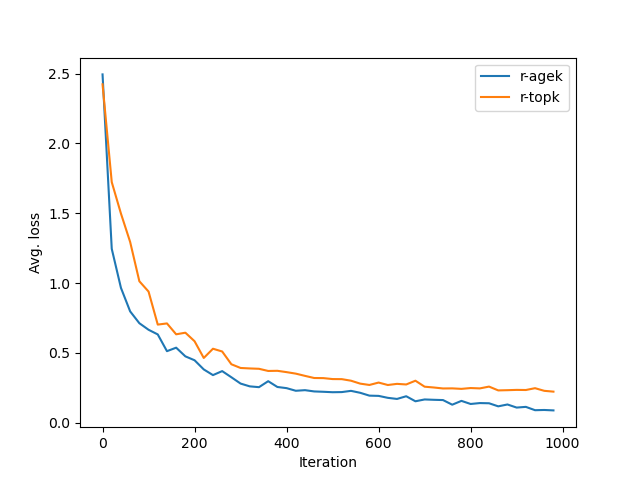}}
    \end{center}
    \caption{(a) Accuracy (in percentage) and (b) loss (averaged over all 10 users) over training iterations with MNIST dataset.}
    \label{MNIST_performance}
\end{figure}

The accuracy performance (in percentage) and the loss corresponding to the two frameworks, with the MNIST dataset, averaged over all 10 users, is plotted in Fig.~\ref{MNIST_performance}(a) and Fig.~\ref{MNIST_performance}(b), respectively. As shown in Fig.~\ref{MNIST_performance}(a), our framework demonstrates faster convergence compared to the \emph{rTop-k} algorithm, attributed to the ability of \emph{rAge-k} to cluster users with similar data types. We also note that \emph{rAge-k} achieves better accuracy performance by the end of the training process. The comparison of loss values achieved by our age-based approach and the \emph{rTop-k} algorithm is shown in Fig.~\ref{MNIST_performance}(b). Our method exhibits a faster reduction in loss compared to the \emph{rTop-k} framework where clients independently transmit sparse updates to the PS in a non-i.i.d.~client data setting. This improvement underscores our framework's capability to address both data heterogeneity and communication overhead issues through accurate clustering of users with similar data distribution via their age vectors at the PS.

\begin{figure}[!h]
    \begin{center}
    \subfigure[]{%
    \includegraphics[width=0.49\linewidth]{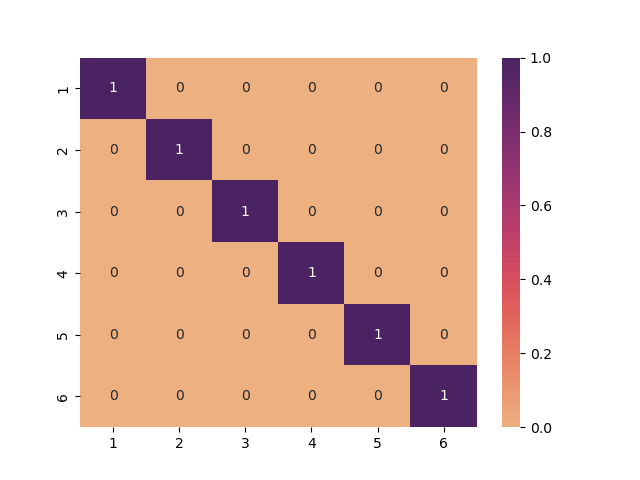}}
    \subfigure[]{%
    \includegraphics[width=0.49\linewidth]{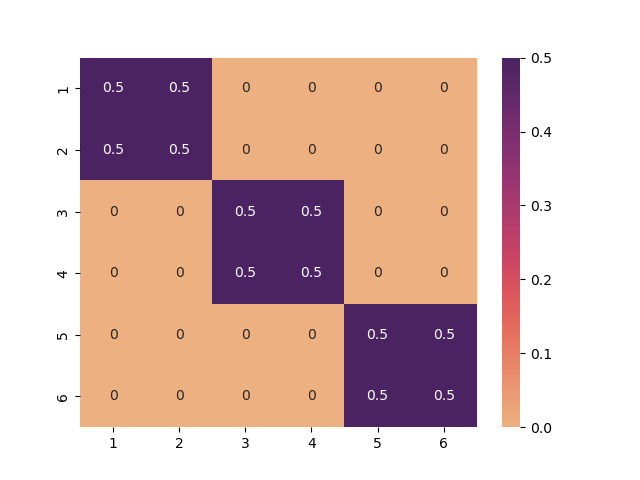}}
    \end{center}
    \caption{Heatmap of the connectivity matrix achieved by the DBSCAN method over different epochs, (a) iteration 1, (b) iteration 201}
    \label{heatmap_synthetic_cifar}
\end{figure}

\begin{figure}[!h]
    \begin{center}
    \subfigure[]{%
    \includegraphics[width=0.49\linewidth]{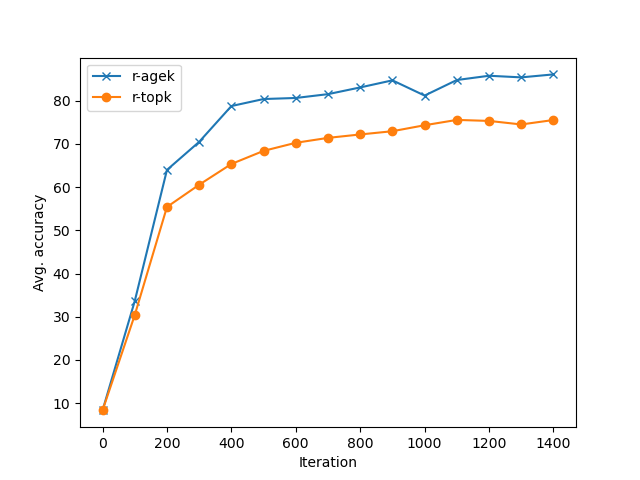}}
    \subfigure[]{%
    \includegraphics[width=0.49\linewidth]{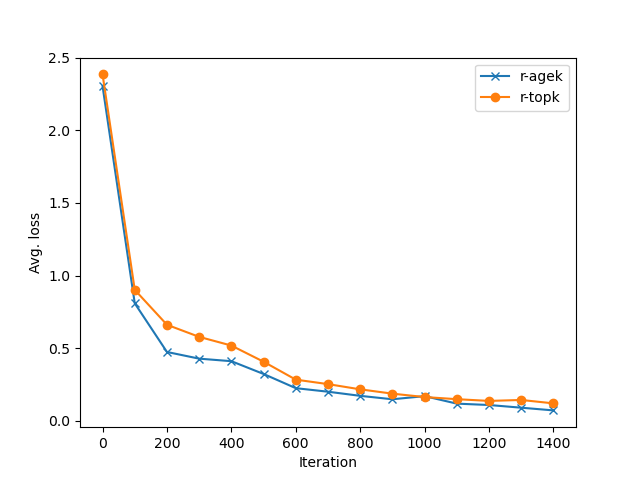}}
    \end{center}
    \caption{(a) Accuracy (in percentage) and (b) loss (averaged over all 6 users) over training iterations with CIFAR10 dataset.}
    \label{CIFAR_performance}
\end{figure}

We similarly evaluate the two frameworks on the CIFAR10 dataset, this time with a total of 6 clients. We distribute the categories amongst clients to create non-i.i.d.~client datasets in a similar fashion to the MNIST case, i.e., we assign statistically similar datasets to pairs of clients. To elaborate on the category assignment among clients, clients $1$ and $2$ handle images with label $1$, $2$, $3$,  clients $3$ and $4$ handle images with label $4$, $5$, $6$, and  clients $5$ and $6$ handle images with label $7$, $8$, $9$, $10$. Fig.~\ref{heatmap_synthetic_cifar} illustrates the clustering effectiveness of DBSCAN, successfully grouping users with similar classes after 200 iterations. The accuracy performance and loss performance for both the frameworks is illustrated in Fig.~\ref{CIFAR_performance}(a) and Fig.~\ref{CIFAR_performance}(b). Our framework provides a significant boost to the accuracy performance for the CIFAR10 dataset, reaching $80\%$ accuracy by iteration $400$, while the \emph{rTop-k} algorithm only achieves $70\%$ accuracy by iteration 1$400$. Similarly, our framework attains lower loss values much sooner compared to the \emph{rTop-k} algorithm as shown in Fig.~\ref{CIFAR_performance}(b).  

\section{Conclusion and Discussion}
We proposed \emph{rAge-k}, a novel communication-efficient FL framework that utilizes the AoI metric to sparsify gradient vectors transmitted from clients to the PS. Our framework not only reduces communication overhead but also effectively addresses data heterogeneity by clustering users with similar data types. Our experimental results showed that the \emph{rAge-k} framework outperforms conventional compression algorithm \emph{rTop-k}, achieving faster and more accurate results under the same bandwidth.

Moving forward, this framework can be extended to enhance model performance on more complex datasets, such as CelebA and ImageNet, by incorporating personalization layers. In this extended setup, all users could collaborate on a shared base model via the PS, while clients within the same cluster could exchange personalized models among themselves. This approach transforms our framework into a form of semi-supervised FL, showcasing its potential for broader applicability and improved model adaptability in heterogeneous data environments.
 
\bibliographystyle{unsrt}
\bibliography{reference}

\begin{thebibliography}{10}

\bibitem{mcmahan2017communication}
B.~McMahan, E.~Moore, D.~Ramage, S.~Hampson, and B.~A. y~Arcas.
\newblock Communication-efficient learning of deep networks from decentralized data.
\newblock In {\em Artificial Intelligence and Statistics}, pages 1273--1282. PMLR, 2017.

\bibitem{hetero_noniid_1}
T.~Li, A.~K. Sahu, M.~Zaheer, M.~Sanjabi, A.~Talwalkar, and V.~Smith.
\newblock Federated optimization in heterogeneous networks.
\newblock In {\em MLSys}, March 2020.

\bibitem{hetro_noniid_2}
S.~P. Karimireddy, S.~Kale, M.~Mohri, S.~Reddi, S.~Stich, and A.~T. Suresh.
\newblock {SCAFFOLD}: Stochastic controlled averaging for federated learning.
\newblock In {\em ICML}, July 2020.

\bibitem{sattler2019robust}
F.~Sattler, S.~Wiedemann, K.~R. M{\"u}ller, and W.~Samek.
\newblock Robust and communication-efficient federated learning from non-iid data.
\newblock {\em IEEE Transactions on Neural Networks and Learning Systems}, 31(9):3400--3413, September 2019.

\bibitem{bernstein2018signsgd}
J.~Bernstein, Y.~X. Wang, K.~Azizzadenesheli, and A.~Anandkumar.
\newblock signsgd: Compressed optimisation for non-convex problems.
\newblock In {\em ICML}, July 2018.

\bibitem{wen2017terngrad}
W.~Wen, C.~Xu, F.~Yan, C.~Wu, Y.~Wang, Y.~Chen, and H.~Li.
\newblock Terngrad: Ternary gradients to reduce communication in distributed deep learning.
\newblock {\em Neurips}, December 2017.

\bibitem{basu2019qsparse}
D.~Basu, D.~Data, C.~Karakus, and S.~Diggavi.
\newblock Qsparse-local-sgd: Distributed sgd with quantization, sparsification and local computations.
\newblock {\em Neurips}, December 2019.

\bibitem{Reisizadeh2020FedPAQAC}
A.~Reisizadeh, A.~Mokhtari, H.~Hassani, A.~Jadbabaie, and R.~Pedarsani.
\newblock Fedpaq: A communication-efficient federated learning method with periodic averaging and quantization.
\newblock In {\em AISTATS}, August 2020.

\bibitem{top_k}
Y.~Lin, S.~Han, H.~Mao, Y.~Wang, and B.~Dally.
\newblock Deep gradient compression: Reducing the communication bandwidth for distributed training.
\newblock In {\em ICLR}, May 2018.

\bibitem{barnes2020rtop}
L.~P. Barnes, H.~A. Inan, Be. Isik, and A.~{\"O}zg{\"u}r.
\newblock {$rTop-k$}: A statistical estimation approach to distributed sgd.
\newblock {\em IEEE Journal on Selected Areas in Information Theory}, 1(3):897--907, November 2020.

\bibitem{stich2018sparsified}
S.~U. Stich, J.~B. Cordonnier, and M.~Jaggi.
\newblock Sparsified sgd with memory.
\newblock {\em Neurips}, December 2018.

\bibitem{isik2023communication}
B.~Isik, F.~Pase, D.~Gunduz, S.~Koyejo, T.~Weissman, and M.~Zorzi.
\newblock Communication-efficient federated learning through importance sampling.
\newblock Available online at arXiv:2306.12625.

\bibitem{arivazhagan2019federated}
M.~G. Arivazhagan, V.~Aggarwal, A.~K. Singh, and S.~Choudhary.
\newblock Federated learning with personalization layers.
\newblock Available online at arXiv:1912.00818.

\bibitem{collins2021exploiting}
L.~Collins, H.~Hassani, A.~Mokhtari, and S.~Shakkottai.
\newblock Exploiting shared representations for personalized federated learning.
\newblock In {\em ICML}, July 2021.

\bibitem{FedGradNorm}
M.~Mortaheb, C.~Vahapoglu, and S.~Ulukus.
\newblock Fedgradnorm: Personalized federated gradient-normalized multi-task learning.
\newblock In {\em IEEE SPAWC}, July 2022.

\bibitem{morafah2022flis}
M.~Morafah, S.~Vahidian, W.~Wang, and B.~Lin.
\newblock Flis: Clustered federated learning via inference similarity for non-iid data distribution.
\newblock Available online at arXiv:2208.09754.

\bibitem{mortaheb2023personalized}
M.~Mortaheb and S.~Ulukus.
\newblock Personalized decentralized multi-task learning over dynamic communication graphs.
\newblock In {\em IEEE CISS}, March 2023.

\bibitem{sattler2020clustered}
F.~Sattler, K.~R. M{\"u}ller, and W.~Samek.
\newblock Clustered federated learning: Model-agnostic distributed multitask optimization under privacy constraints.
\newblock {\em IEEE Transactions on Neural Networks and Learning Systems}, 32(8):3710--3722, August 2020.

\bibitem{yates2021age}
R.~D. Yates, Y.~Sun, D.~R. Brown, S.~K. Kaul, E.~Modiano, and S.~Ulukus.
\newblock Age of information: An introduction and survey.
\newblock {\em IEEE Journal on Selected Areas in Communications}, 39(5):1183--1210, May 2021.

\bibitem{yang2020age}
H.~H. Yang, A.~Arafa, T.~Q. Quek, and H.~V. Poor.
\newblock Age-based scheduling policy for federated learning in mobile edge networks.
\newblock In {\em IEEE ICASSP}, May 2020.

\bibitem{ozfatura2020age}
E.~Ozfatura, B.~Buyukates, D.~G{\"u}nd{\"u}z, and S.~Ulukus.
\newblock Age-based coded computation for bias reduction in distributed learning.
\newblock In {\em IEEE Globecom}, December 2020.

\bibitem{buyukates2020timely}
B.~Buyukates and S.~Ulukus.
\newblock Timely distributed computation with stragglers.
\newblock {\em IEEE Transactions on Communications}, 68(9):5273--5282, September 2020.

\bibitem{wang2022age}
K.~Wang, Y.~Ma, M.~B. Mashhadi, C.~H. Foh, R.~Tafazolli, and Z.~Ding.
\newblock Age of information in federated learning over wireless networks.
\newblock Available online at arXiv:2209.06623.

\bibitem{buyukates2021timely}
B.~Buyukates and S.~Ulukus.
\newblock Timely communication in federated learning.
\newblock In {\em IEEE Infocom}, May 2021.

\bibitem{ester1996density}
M.~Ester, H.~Kriegel, J.~Sander, X.~Xu, et~al.
\newblock A density-based algorithm for discovering clusters in large spatial databases with noise.
\newblock In {\em KDD}, volume~96, pages 226--231, 1996.

\end{thebibliography}

\end{document}